\documentclass{article}
\usepackage[utf8]{inputenc}

\title{\textbf{Molecular generative Graph Neural Networks for Drug Discovery}}
\author{Pietro Bongini\textsuperscript{a,b,c}, Monica Bianchini\textsuperscript{a}, Franco Scarselli\textsuperscript{a}}

\usepackage{subcaption}
\usepackage{url}
\usepackage{graphicx}
\usepackage{makeidx}
\usepackage{color}
\usepackage{soul}
\usepackage{amsmath}
\usepackage{algorithm}
\usepackage{algpseudocode}

\begin{document}

\maketitle

\noindent
\textsuperscript{a} \small{University of Siena - Department of Information Engineering and Mathematics - via Roma 56, 53100, Siena, Italy}
\\
\\
\textsuperscript{b} \small{University of Florence - Department of Information Engineering - via di Santa Marta 3, 50139, Florence, Italy}
\\
\\
\textsuperscript{c} \small{Corresponding author, email: pietro.bongini@unifi.it}\\

\begin{abstract}
\noindent
Drug Discovery is a fundamental and ever-evolving field of research. The design of new candidate molecules requires large amounts of time and money, and computational methods are being increasingly employed to cut these costs. Machine learning methods are ideal for the design of large amounts of potential new candidate molecules, which are naturally represented as graphs. Graph generation is being revolutionized by deep learning methods, and molecular generation is one of its most promising applications. In this paper, we introduce a sequential molecular graph generator based on a set of graph neural network modules, which we call MG\textsuperscript{2}N\textsuperscript{2}. At each step, a node or a group of nodes is added to the graph, along with its connections. The modular architecture simplifies the training procedure, also allowing an independent retraining of a single module. Sequentiality and modularity make the generation process interpretable. The use of graph neural networks maximizes the information in input at each generative step, which consists of the subgraph produced during the previous steps.
Experiments of unconditional generation on the QM9 and Zinc datasets show that our model is capable of generalizing molecular patterns seen during the training phase, without overfitting. The results indicate that our method is competitive, and outperforms challenging baselines for unconditional generation.
\end{abstract}

\section*{Keywords}
\noindent
Graph Generation, Molecule Generation, Deep Learning, Graph Neural Networks, Drug Discovery.

\newpage

%% main text
\section{Introduction}
\label{Introduction}
\noindent
Graph generation is a complex problem with several real--world applications, and many different approaches have been devised for solving it. Classical methods resorted to random mathematical models: The Erd\"{o}s--R{\'e}nyi model \cite{ErdosRenyi} was the first approach in this direction. Since Erd\"{o}s--R{\'e}nyi graphs tend to have unrealistically low clustering coefficients, especially with respect to community graphs, two methods were later developed, mainly to obtain more realistic small--world networks: the growth--based Barab\'asi--Albert model \cite{BarabasiAlbert}, and the Watts--Strogatz rewiring model \cite{WattsStrogatz}. The recent developments in machine learning have stimulated its application to the field of graph generation. Deep learning techniques, indeed, can capture the characteristics of a given domain from a set of examples, which are then exploited to generate new graphs. Variational Auto--Encoders (VAEs) \cite{VAE_original} were the first neural network models to be employed for this purpose \cite{VGAE} \cite{Graphite}.  The success of Generative Adversarial Networks (GANs) \cite{GAN_original} in image generation has led to replicate the same adversarial approach for graph--structured data \cite{GraphGAN} \cite{NetGAN}. This approach can be improved by adding constraints to the adversarial learning \cite{ConstrainedGraphGAN}. The different nature of the problem, though, has encouraged the development of alternative solutions as well. While VAEs, by sampling representations from a continuous latent space, can generate graphs as unitary entities, many methods tackle the problem with a sequential approach. The construction of a graph becomes, therefore, a sequence of decisions, in which a node or a group of nodes is added to the graph at each step. On the one hand, many methods make use of Recurrent Neural Networks (RNNs) to handle the decision sequence \cite{GraphRNN}, \cite{GRAN}, \cite{edge_based_sequential_graph_generation}). On the other hand, Graph Neural Networks (GNNs) \cite{GNN_Model}, with their capability of processing graph--structured data without loss of connectivity information, allow to build very powerful generative models. In particular, at each step, GNNs can exploit all the information contained in the partial graph generated by the previous steps, while recurrent models typically rely only on the sequence of previous decisions. In principle, this holds true for any GNN model, including Graph Nets \cite{GraphNets}, GraphSAGE \cite{GraphSAGE}, Gated Graph Sequence Neural Networks \cite{GGSNN}, Message Passing Neural Networks \cite{MPNN}, and Graph Convolution Networks \cite{GCN_spectral_Bruna} \cite{GCN_standard} \cite{GCN_spectral_Defferrard}. The only approach of this type we are aware of is DeepGMG \cite{DeepGMG}, based on Graph Nets \cite{GraphNets}.
\\
The generation of molecular graphs is a complex task, which can lead to the development of new instruments for drug discovery, potentially cutting the huge costs, in terms of both time and money, of that fundamental research process \cite{drug_discovery_costs}. The space of molecular graphs is virtually infinite, and even constraining the size of molecules to few atoms, the number of theoretically possible compounds is overwhelming. Efficient automatic generative techniques are required for the exploration of such huge space, and deep generative models represent ideal candidates. Using SMILES notation \cite{SMILES}, molecules can be generated as sequential objects. This approach has been carried out with VAE models \cite{GomezBombarelli2018}, also exploiting the grammar of the SMILES language \cite{GrammarVAE} \cite{SyntaxVAE}. However, SMILES strings do not preserve the full connectivity information, as molecules are more naturally represented as undirected graphs, with finite (and relatively small) sets of vertex and edge types. Graph--based VAEs have been employed in the generation of molecular graphs \cite{GraphVAE}. Junction--Tree VAEs build graphs by connecting pre--extracted structural motifs \cite{JTVAE}, an approach which has been recently extended to larger molecules and polymers by making the VAE hierarchical \cite{hierarchical_molecule_generation}. This approach can also be improved by exploiting the valence histogram of each atom \cite{CCGVAE}. Recently, statistical flow models, characterized by an invertible encoder/decoder, have been developed \cite{GraphFlow_autoregressive}. Graph GANs have been employed for the generation of molecules, handling the decisions with Reinforcement Learning (RL) techniques \cite{MolGAN} \cite{GCPN}. Finally, the above mentioned approach based on Graph Nets have also been applied to this task with promising results \cite{DeepGMG}.
\\
In this work, we present a sequential molecular graph generator based on GNNs \cite{GNN_Model}, which we call Molecule Generative Graph Neural Network (MG\textsuperscript{2}N\textsuperscript{2}). A single node is added and connected to the graph, at each step. The method focuses on one node at a time, and generates its neighbors before processing the following node, preventing disconnected components from being created. Similarly to GraphRNN \cite{GraphRNN}, we follow a Breadth First Search (BFS) ordering to decide which nodes to expand first. Edges are generated in parallel rather than sequentially, making the approach less computationally demanding. The control flow in MG\textsuperscript{2}N\textsuperscript{2} depends on decisions implemented by three GNN modules. The sequential and modular nature of our method makes it interpretable. As previously mentioned, at each step, GNNs exploit all the information contained in the subgraph generated until that step. Gumbel softmax \cite{gumbel_softmax} output layers allow the networks to be trained over discrete stochastic distributions. Moreover, the modules are trained independently of each other: This feature simplifies the learning process and allows to retrain each module independently. The GNN model used in this work was derived from the original GNN approach \cite{GNN_Model}, which was proved to be a universal approximator on graphs \cite{GNN_universal_approximation}. 
This property ensures that the GNN model is general enough to be able to make the complex decisions that the modules must implement.
\\
The contributions of this work consist in a new sequential generative model for molecular graphs, MG\textsuperscript{2}N\textsuperscript{2}, and its experimental evaluation on two well--known benchmarks for the generation of small organic molecules, the Quantum Machine 9 (QM9) and Zinc datasets. The main novelty of our approach consists in using the GNN framework for molecular graph generation, with a modular architecture, in order to maximize the information and make the generative model flexible. The results show that the proposed approach outperforms very competitive baselines in the task of unconditional generation. The experiments also clarify the main properties of the method and show that
MG\textsuperscript{2}N\textsuperscript{2} is capable of generating molecules with chemical characteristics similar to those of the original datasets.
\\
The rest of this paper is organized as follows. A short description of the GNN model used in this work is provided in Section \ref{model}. Section \ref{Method} presents and discusses the generative algorithm and its implementation with neural networks. The experiments and their results are described and commented in Section \ref{Experiments}. Our conclusions are drawn in Section \ref{Conclusions}.

\section{The GNN Model}
\label{model}
\noindent
The Graph Neural Networks used in this paper are based on the original work \cite{GNN_Model}, to which we redirect for a detailed mathematical formulation of the model. The implementation is described in the following. Each graph $G = (V, E)$ is composed of a set of vertices $V$ and a set of edges $E \subseteq V \times V $. Each vertex $v_{i} \in V$ is labeled with a vector of attributes $l_{i}$. Each edge $(v_{i}, v_{j}) \in E$ is labeled with a vector of attributes $e_{i,j}$. Based on $E$, a neighborhood function, assigning a set of neighbors $Ne(v_{i})$ to each vertex $v_{i} \in V$ is defined as $Ne(v_{i}) = \{v_{j} : (v_{j},v_{i}) \in E\}$.
\\
A state $s_{i}$ is assigned to each vertex $v_{i} \in V$. It is initialized as $s_{i}^{0} = l_{i}$. The state of all the vertices is then updated iteratively, and in parallel, until convergence and/or a maximum number of iterations are reached. During this process, each vertex communicates with its neighbors by sending and receiving messages. The state $s_{i}^{k}$ of vertex $v_{i}$, at step $k$, is calculated as in Eq. (\ref{eq:state_update}), based on its state at the previous step $s_{i}^{k-1}$ and the messages coming from its neighbors $v_{j} \in Ne(v_{i})$.
\begin{equation}
	\label{eq:state_update}
	s_{i}^{k} = F(s_{i}^{k-1},\ A(\{M(v_{i}, v_{j}, k): v_{j} \in Ne(v_{i}) \}))
\end{equation}
\noindent
In Eq. (\ref{eq:state_update}), some message passing function $M$ defines the message sent from $v_{j}$ to $v_{i}$ at step $k$. A neighborhood aggregation function $A$ defines how the messages coming from all the neighbors of $v_{i}$ are aggregated. Finally, a state updating function $F$ defines how the new state is calculated. In principle, all of these functions could be learnt by neural networks.
In earlier models \cite{GNN_Implementation}, $M$ was implemented by a MultiLayer Perceptron (MLP), while the other two functions were kept fixed. In this work, we opted for learning $F$ with an MLP, while keeping $M$ and $A$ fixed. This choice is also in line with other recent developments, which are broadly summarized in \cite{GraphNets}\cite{GraphSAGE}. In particular, we defined $M$ as the concatenation of the state of the source node $s_{j}^{k-1}$ and the label $e_{i,j}$ of the edge $(v_{j},v_{i})$, as shown in Eq. (\ref{eq:message}).
\begin{equation}
	\label{eq:message}
	M(v_{i}, v_{j}, k) = ( s_{j}^{k-1},\ e_{i,j} )
\end{equation}
\noindent
Two different neighborhood aggregation functions were employed, defined as either the arithmetic mean or the element--wise sum of the incoming messages, as shown in (\ref{eq:aggregation}).
\begin{align}
	\begin{aligned}
		\label{eq:aggregation}
		A_{avg}( \{ M (v_{i}, v_{j}, k) : v_{j} \in Ne(v_{i}) \} ) &= \frac{1}{|Ne(v_{i})|}\sum_{j : v_{j} \in Ne(v_{i})} M(v_{i}, v_{j}, k)
		\\
		A_{sum}( \{ M (v_{i}, v_{j}, k) : v_{j} \in Ne(v_{i}) \} ) &= \sum_{j : v_{j} \in Ne(v_{i})} M(v_{i}, v_{j}, k)
	\end{aligned}
\end{align}
\noindent
Therefore, we can introduce a hyperparameter coefficient $a$ which allows us to choose between the aggregation functions $A_{avg}$ and $A_{sum}$, by simply assigning it the values $1/|Ne(v_{i})|$ or $1$, respectively. As a consequence, the final form of (\ref{eq:state_update}) can be written as in (\ref{eq:state_update_final}).
\begin{equation}
	\label{eq:state_update_final}
	s_{i}^{k} = F ( s_{i}^{k-1},\ a \sum_{j : v_{j} \in Ne(v_{i})} ( s_{j}^{k-1},\ e_{i,j} ) )
\end{equation}
\noindent
The state updating process is stopped when either state convergence or a maximum number of iterations $k = k_{max}$ are reached. State convergence is assumed to happen when the distance between the current state and the previous state becomes negligible: $\|s_{i}^{k} - s_{i}^{k-1}\| < \epsilon$. $\epsilon$ and $k_{max}$ are set as hyperparameters, $\|\cdot\|$ is the Euclidean norm. 
\\
The output $y \in \mathbf{R}^{p}$ is then determined by an output function $O$ which depends on the type of problem at hand. In graph--based problems it is defined on the whole graph $G$, in node--based problems it is defined on single vertices $v_{i} \in V_{out} \subseteq V$, while in edge--based problems it is defined on single edges $(v_{i}, v_{j}) \in E_{out} \subseteq E$, where $V_{out}$ and $E_{out}$ are respectively the subset of vertices and the subset of edges for which an output is defined. 
\\
In this work, the problem of graph generation is divided into three classification subtasks. One of them is node--based while the other two are edge--based. In the former, the output function is defined as in Eq. (\ref{eq:node_output}), while in the latter, the output function is defined as in Eq. (\ref{eq:edge_output}). 
\begin{equation}
	\label{eq:node_output}
	y_{i} = O(s_{i}^{k^{*}}) 
\end{equation}
\begin{equation}
	\label{eq:edge_output}
	y_{i,j} = O(s_{i}^{k^{*}}, s_{j}^{k^{*}}, e_{i,j}) 
\end{equation}
\noindent
In both equations, $k^{*}$ represents the last iteration of Eq. (\ref{eq:state_update_final}), determined by one or both of the stopping conditions. The function $O$ is implemented by an MLP, which is put in cascade to the state network. Learning is achieved by comparing, through a loss function $E$, the supervision $y$ to the output $\hat{y}$ produced by the output network. The network weights are updated by Backpropagation Through Time \cite{backpropagation_through_time}.

\section{Method}
\label{Method}
\noindent
Our method consists in a graph generation algorithm tailored to the production of small organic molecules, and its implementation with GNNs. Subsection \ref{Method:algorithm} describes the algorithm, while the implementation is presented in Subsection \ref{Method:models}. The details about preprocessing are contained in Subsection \ref{Method:preprocessing}, while Subsection \ref{Method:ordering} deals with node ordering issues.

\subsection{Generation Algorithm}
\label{Method:algorithm}
\noindent
The generation of a labeled graph $G=(V,E)$ is handled as a sequential process, starting with an initially empty $E$ and with a single vertex $V = \{v_{0}\}$. The label $l_{0}$ of $v_{0}$ is sampled from a distribution of labels $D_{0}$, which is learned from the training set. Each step consists in adding a new node to the graph and connecting it to the other nodes. The algorithm focuses on one node $v_{i}$ at a time, generating all its neighbors before focusing on the following node $i=i+1$. This process will be referred to as \emph{node expansion}. Nodes are indexed according to the order in which they have been generated, so that, for instance, the third generated node $v_{3}$ will always be the fourth node to be expanded ($v_{0}$ is the first). The process stops when all the nodes have been expanded ($i > |V|$) or when the maximum number of nodes has been reached ($|V| = |V_{max}|$).
\\
As a new node $v_j$ is generated, first it is connected to the node $v_i$ which is being expanded, then it can be linked to the other vertices $V \setminus \{v_{i}, v_{j}\}$. While the set of edges generated in the latter phase can be empty, the $(v_{i}, v_{j})$ edge is always generated. This constraint ensures that the generated graph is always connected, without impairing generality: any graph can still be produced. 
\\
We can define three problems that must be solved to carry out a generative step. Each problem corresponds to a function the model will provide: node generation ($P1$), first edge classification ($P2$), additional node linking ($P3$).
\begin{itemize}
	\item $P1$ decides whether to expand $v_{i}$ with a new neighbor node $v_{j}$ or to stop its expansion. If $v_{j}$ is generated, $P1$ also returns its label $l_{j}$.
	\item $P2$ is called after a new node $v_{j}$ has been generated. It determines the label $e_{i,j}$ of the edge $(v_{i},v_{j})$.
	\item $P3$ is called after a new node $v_{j}$ has been generated and connected to $v_i$. It determines the existence of any possible edge connecting $v_{j}$ to any other vertex $v_{k} \in V \setminus \{v_{i},v_{j}\}$. The labels of all the generated edges are also returned. All the edges are processed in parallel. The main drawback of this approach is that the dependencies between edges are ignored, but it also brings the advantages of avoiding edge ordering biases and of significantly reducing the time cost.
\end{itemize} 
The generation algorithm is summarized in Algorithm \ref{alg:generation}.

\begin{algorithm}
\caption{Graph generation algorithm.}
\label{alg:generation}
\begin{algorithmic}
\Procedure{Generate}{$G = (V,E)$}
\State $V\gets \{v_{0}\}, l_{0} \sim D_{0}$
\State $E\gets \emptyset$
\State $i\gets 0$
\State $j\gets 1$
\While{$(i < |V|) \land (|V| \leq |V|_{max})$}\label{while:node}
\State $gd \gets P1(V,E,i)$
\While{$gd \neq stop$}\label{while:step}
\State $V\gets V \cup \{v_{j}\}, l_{j} \gets gd$
\State $E\gets E \cup \{(v_{i}, v_{j})\}$
\State $e_{i,j}\gets P2(V,E,i,j)$
\For{$k \in [0,j-1], k \neq i$}\label{for:edges} \Comment{Parallel Execution}
\State $ld \gets P3(V,E,k,j)$
\If{$ld \neq disconnected$}\label{if:link}
\State $E \gets E \cup \{(v_{k},v_{j})\}$
\State $e_{k,j} \gets ld$
\EndIf\label{endif:link}
\EndFor\label{endfor:edges}
\State $j\gets j + 1$
\State $gd \gets GeneratorDecision(V,E,i)$
\EndWhile\label{endwhile:step}
\State $i\gets i + 1$
\EndWhile\label{endwhile:node}
\State \textbf{return} $G = (V,E)$
\EndProcedure
\end{algorithmic}
\end{algorithm}

\subsection{Implementation with Graph Neural Networks}
\label{Method:models}
\begin{figure}[!htb]
	\centering
	\begin{minipage}{0.98\textwidth}
		\includegraphics[width=\linewidth]{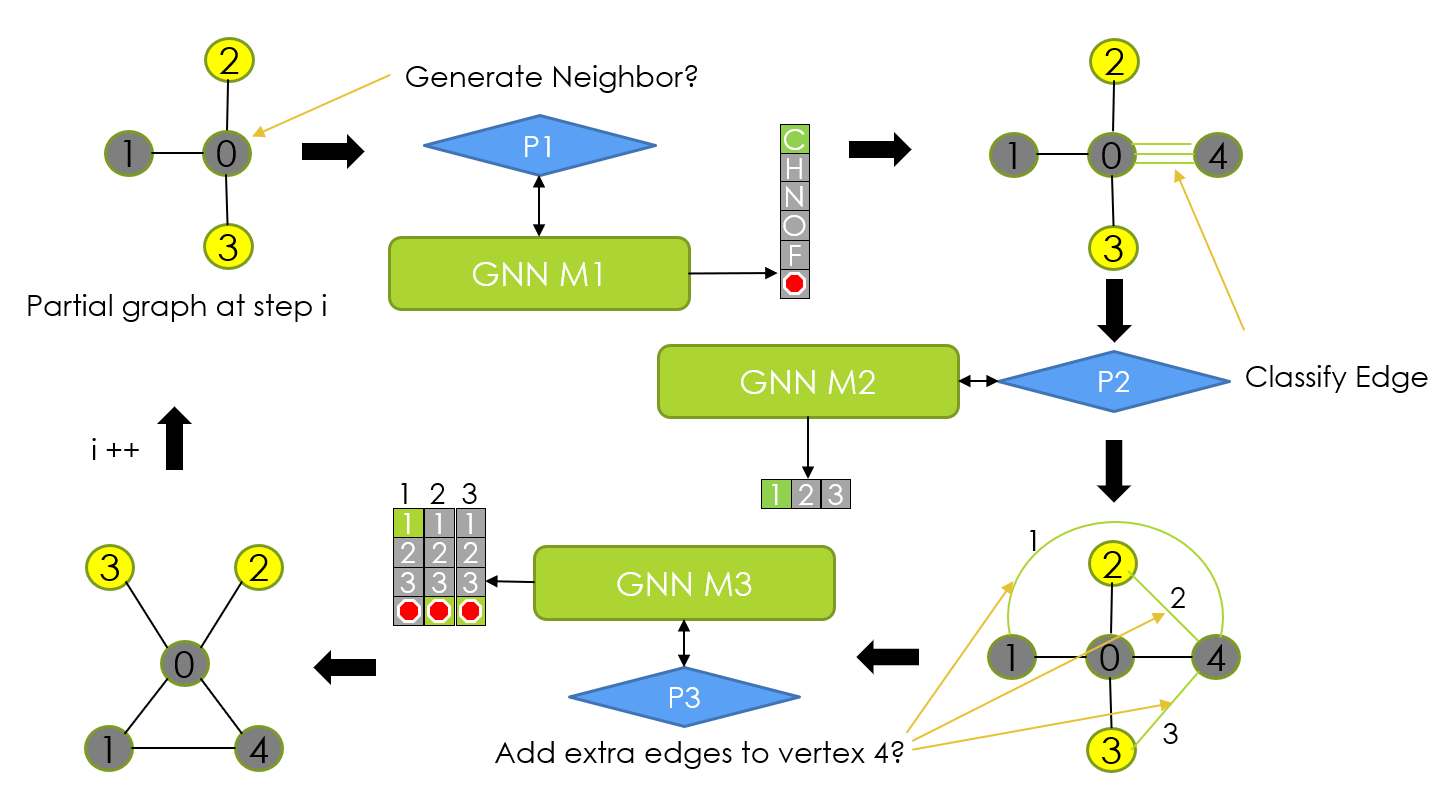}
	\end{minipage}\hfill
	\caption{\small{Flowchart of the generation algorithm. An example step is summarized, with the three GNN modules (M1, M2, M3), the three problems they are assigned to (P1, P2, P3), their inputs and their outputs. Grey nodes represent carbon atoms, while yellow nodes represent hydrogen atoms. Green edges stand for candidate edges, while black edges represent single bonds. C,H,N,O,F are the element symbols. Classes 1,2,3 represent single, double, and triple bonds, respectively. Red octagons stand for the \textit{stop} decision (M1) or the \textit{do not generate this edge} decision (M3)}}
	\label{fig:flowchart}
\end{figure}

\noindent
In the proposed approach, each of the functions $P1$, $P2$, $P3$ described in Subsection \ref{Method:algorithm} is implemented by a dedicated GNN module, which will be referred to as M1, M2, M3, respectively. Each of the modules is trained separately, and one step at a time, assuming the other two modules' decisions to always correspond to the ground truth. This is a strong assumption, which will prevent the model from exploring possible different solutions, but it dramatically simplifies the training procedure. Another advantage of this paradigm is the fact that, each being trained separately from the others, the modules can be recombined to build new versions of the model. If a module needs to be optimized there is no need of re--training the other two.
\\
In order to generate labeled graphs, we need to make some assumptions on the nature of vertex and edge labels. Three main cases can be identified: unlabeled graphs, graphs with continuous vertex and edge labels, graphs with a finite set of vertex and edge types. In this work, we will focus on the third case, which corresponds to the typical setting in molecule generation problems. Thus, in the following, we assume that the label $l_{i}$ of any vertex $v_{i}$ belongs to a finite set $l_{i} \in T_{v}$, the label $e_{i,j}$ of any edge $(v_{i},v_{j})$ belongs to a finite set of types $e_{i,j} \in T_{e}$, and $T_{v}$ and $T_{e}$ are defined by the dataset. 
\\
In the proposed approach, the GNN modules generate nodes and edges along with their labels. With reference to Algorithm \ref{alg:generation}, the following holds.
\begin{itemize}
	\item $P1$ faces a node--based classification problem, as it decides whether to stop the expansion of the current node $v_{i}$ or to generate another neighbor $v_{j}$, and, in case, which label to assign to $v_{j}$. The set of output vertices of M1 consists only of $v_{i}$: $V_{out} = \{v_{i}\}$. The output classes correspond to the union of the $stop$ decision to the set of vertex types $\{stop\} \cup T_{v}$.
	\item $P2$ deals with an edge--based classification problem, since it generates the label of the edge connecting the vertex being expanded $v_{i}$ and its new neighbor $v_{j}$. The set of output edges of M2 consists only of this edge $E_{out} = \{(v_{i},v_{j})\}$. The output classes correspond to the set of edge types $T_{e}$.
	\item $P3$ works on an edge--based classification problem, since it predicts the existence, and, in case, the label, of every possible edge connecting the new vertex $v_{j}$ to the other nodes in the graph, except the node being expanded $v_{i}$. These calls are parallelized and integrated in a single prediction from M3. This idea has the drawback of considering each predictable edge as if it were independent from the other predictable edges, but it also allows to avoid the biases introduced by taking the decisions in sequence and it speeds up the procedure. To do so, the graph $G$ is extended with a set of provisional edges $E_{p} = \{(v_{k},v_{j}) : v_{k} \in V \setminus \{v_{i},v_{j}\}\}$. The module M3 takes in input the new graph $G'=(V,E'): E' = E \cup E_{p}$. The set of output edges for M3 is $E_{out}'=E_{p}$. The output classes correspond to the union of the $disconnected$ decision to the set of edge types $\{disconnected\} \cup T_{e}$.
\end{itemize}
An example step of this algorithm is visually summarized as a flowchart in Figure \ref{fig:flowchart}.

\noindent
To learn a stochastic behavior from the supervisions, which are samples from a categorical distribution, we resorted to a Gumbel softmax output layer \cite{gumbel_softmax}, based on the Gumbel--Max method for sampling from discrete distributions \cite{gumbel_original} \cite{gumbel_a_star}.
\\
This approach allows to backpropagate through an arbitrarily close approximation of the categorical distribution. The softmax can be annealed, by decreasing a temperature parameter $\tau$, from a less accurate approximation, which tends to a uniform distribution for $\tau \rightarrow \infty$, and to the discrete distribution itself for $\tau \rightarrow 0$. Lower temperatures come at the cost of an increasing gradient variance. The choice of two parameters $\tau_{max}$ and $\tau_{min}$, and a curve, will determine the annealing path. Annealing while training has the positive effect of encouraging the exploration of alternatives to the decision with the highest estimated probability in the early phases, to then converge to more canonical decisions in the final training epochs, when the estimation of the class probabilities has gained higher reliability. This is very important to prevent the networks from learning repetitive patterns, and to avoid mode collapse (i.e. generating always vertices of the same type, based on the highest prior probability).

\subsection{Graph Preprocessing}
\label{Method:preprocessing}
\noindent
To build the training, validation, and test sets for M1, M2, M3, the molecules from the dataset under analysis are pre--processed. For each generative step, we need an input graph, the index of the focus node, and a supervision. Each molecular graph $G=(V,E)$ is decomposed in a sequence of incomplete graphs, one for each generative step. 
\\
For M1, the sequence is composed of $n = 2|V|-1$ graphs. The first graph contains only one node $G_{0}=(V_{0} = \{v_{0}\}, E_{0} = \{\})$, any intermediate graph $G_{i}=(V_{i}, E_{i})$ corresponds to an incomplete subgraph of $G$, $G_{i} = ( V_{i} \subset V, E_{i} \subset E )$, and the last graph is complete $G_{n-1} = G$. For M2 and M3, the sequences are composed of $n = |V|-1$ graphs, because M2 and M3 are not called after the $|V|$ $stop$ decisions from M1 (see Algorithm \ref{alg:generation}). The graphs $G_{i} = (V_{i} \subset V, E_{i} \subset E)$ acquire nodes and edges as $i$ grows.
\\
The sets are built so that graphs from the same generative sequence (which correspond to different levels of completion of the same original graph) belong to the same set (and to the same batch). This is particularly important to avoid evaluation biases deriving from testing or validating on examples which have slightly different replicas in the training set. 

\subsection{Node Ordering}
\label{Method:ordering}
\noindent
To define the generative sequences of the graphs, a node ordering needs to be established. This will determine the order in which the nodes of each graph must be generated, and, consequently, the sequences of input graphs and supervisions described in Subsection \ref{Method:preprocessing}. The model is expected to learn this generative strategy from the training set, so that, for instance, a training set in which carbon atoms have higher priority will teach the model to generate carbon neighbors first. Theoretically, being $V$ a set with no given ordering, the model would benefit from being trained on any possible node ordering. Since this is impossible from a computational point of view, some constraints must be imposed to reduce the number of orderings from $o(|V|!)$ to a computationally feasible level. In this work we chose a Breadth First Search (BFS) strategy, which has the additional benefit of reducing the number of link predictions needed at each step \cite{GraphRNN}. Among the nodes at the same depth level in the BFS tree, node types with lower average centrality are expanded first. The average centrality of node types is measured on the training set, according to the Freeman Betweenness Centrality \cite{freeman_betweenness_centrality}. This boosts both positive effects of the BFS strategy. To further reduce the number of possible orderings of a factor $|V|$, we decided to always start from the same node, which is node $0$ of the original node numbering taken from the dataset. The other nodes are then re--numbered according to the previous rules, making a random choice in any case in which multiple permutations are still possible. The latter two assumptions allow us to retain one unique ordering, coming at the cost of a loss of generality. Although this cost would likely be critical for a truly recurrent model, it is sustainable in this learning framework, in which the correlation between two steps is limited to the output of the first shaping the input of the second. The only input to the model, in fact, is represented by the graph itself, regardless to the many possible sequences of steps that may have brought to its current shape.

\section{Experimental Methodology}
\label{Experiments}
\noindent
We performed a series of experiments, testing our model on the QM9 \cite{dataset_QM9}, and Zinc \cite{dataset_Zinc} datasets,  two common benchmarks for the generation of graphs representing small organic molecules, which are introduced in Subsection \ref{Experiments:dataset}. Subsection \ref{Experiments:setup} describes the model setup, deals with the hyperparameters, and introduces the experiments we performed. The evaluation methods are described in Subsection \ref{Experiments:evaluation}, while Subsection \ref{Experiments:results} contains the results, the comparisons with other models, and the considerations on their significance.

\subsection{Dataset Description}
\label{Experiments:dataset}
\noindent
To evaluate our method, a set of experiments were run on the Quantum Machine 9 (QM9) dataset \cite{dataset_QM9}, which is a subset of GDB--17, a chemical universe of 166 billion molecules \cite{dataset_GDB17}. QM9 is an ideal benchmark for a new generative model for molecular graphs, as most competitive methods in this area have been tested on this dataset. It is composed of 133,885 compounds, made of up to 9 heavy atoms (C,O,N,F), plus the hydrogens which are bound to them, for a maximum size of 29 atoms. Each molecule is represented as an undirected graph, in which each vertex corresponds to an atom and each edge corresponds to a chemical bond. The label of each vertex represents the corresponding atom type, through one--hot encoding, so that $|T_{v}|=5$. The label of each edge represents, through one--hot encoding, the type of chemical bond connecting its two vertices, which can be either single, double or triple, so that $|T_{e}|=3$. The output of the modules M1, M2 and M3, defined in Subsection \ref{Method:models} have dimensions, respectively $|\{stop\} \cup T_{v}| = 6$, $|T_{e}| = 3$, and $|\{disconnected\} \cup T_{e}| = 4$.
\\
A random splitting procedure is applied to the dataset, in order to obtain a training set, a test set, and a validation set, composed of 120,000, 10,000 and 3,885 molecular graphs, respectively. The validation set is used, during the training phase, to evaluate the performance of our models on data that are not provided to them directly. The held--out test set allows us to compare the statistics of our sets of generated graphs to the statistics of ground--truth graphs which have never been seen by our generative model, assessing the capability of the model of reproducing the chemical characteristics of QM9 compounds.
\\
Inside each graph, the nodes are re--numbered according to the procedure described in Subsection \ref{Method:ordering}. To determine the order among the neighbors $Ne(v_{i})$ of a generic $v_{i} \in V$, the average Freeman Betweenness Centrality is measured on the 120,000 training graphs, obtaining the following values: FBC(Hydrogen) = 0.0, FBC(Fluorine) = 0.0, FBC(Oxygen) = 0.115, FBC(Nitrogen) = 0.246, FBC(Carbon) = 0.382.
\\
For a further assessment of the generative performance of our model, a second set of experiments is carried out on the Zinc \cite{dataset_Zinc} dataset. This is composed of 249,455 organic molecules of up to 38 heavy atoms (C,O,N,F,P,S,Cl,I,Br). Ring bonds are explicitly labeled as aromatic when part of an aromatic ring. As a consequence, in this setup, we have $|T_{v}|=9$, and $|T_{e}|=4$. The dataset is split into a training set, a test set, and a validation set of 230,000, 10,000, 9,455 molecular graphs, respectively. The training/validation/test procedure is the same described for QM9. The nodes in each single molecular graph are also re--numbered with the same algorithm.

\subsection{Experimental Setup}
\label{Experiments:setup}
\noindent
The code for training the GNNs \cite{GNN_Code} and generating graphs \footnote{Code available at: https://github.com/PietroMSB/MG2N2} was implemented using Tensorflow \cite{tensorflow}. The experiments on QM9, were carried out in the following setup. All the training runs of module M1 were issued on a Nvidia Tesla--V100 GPU, with 32 GB dedicated memory. Training runs of modules M2 and M3 always took place on a Nvidia 2080--Ti GPU. The training set was randomly split in 20 batches of 6,000 graphs each, to reduce the memory requirements. All the experiments used the same split. During the generation of new graphs, even though all the three modules are kept in memory, far less computational resources are needed. The generation sessions were run on the Nvidia 2080--Ti GPU, but required only 0.5 GB of memory. The experiments on Zinc were run on two Nvidia Titan-RTX GPUs, each with 24 GB dedicated memory. The training set was randomly split into 100 batches of 2,300 graphs each to fit in memory.
\\
Table \ref{table:hyperparameters} shows the configurations of the modules M1, M2, M3 used in the QM9 experiments, which include the neighbor aggregation function, the training epochs, the initial learning rate, the maximum number of iterations for state convergence, and the number of hidden units of the state network and the output network. Each GNN module is composed of a state and an output network. The former  is a two--layered MLP implementing the state updating function described in Eq. (\ref{eq:state_update_final}). The latter  is another two--layered MLP, implementing Eq. (\ref{eq:node_output}) in M1, and Eq. (\ref{eq:edge_output}) in M2 and M3. 
The initial values M1(I), M2(I) and M3(I) in Table \ref{table:hyperparameters} were obtained through a preliminary experimentation, with the goal of maximizing the accuracy of the modules M1, M2, M3, each one independently from the others, on the validation set. Just as if the modules had been classifiers, accuracy was calculated as the percentage of correct outputs, according to the single step supervision, and regardless of molecule validity.
\\
For the Gumbel softmax annealing path, based on \cite{gumbel_softmax}, we initially chose a linear descent from $\tau_{max} = 5.0$ to $\tau_{min} = 1.0$ during training. Tests on different linear configurations did not bring improvements. In particular, annealing to temperatures $\tau < 1.0$ brought the model to an unwanted repetitive behavior. Therefore, we kept the initial annealing path for all the successive training runs. All the models were trained with an Adam optimizer \cite{Adam} and cross--entropy loss, which does not require adjustments to work with the Gumbel softmax output layers.

\begin{table}
	\centering
	\scalebox{0.75}{
		\begin{tabular}{ccccccc}
			\hline
			\textbf{Module} & \textbf{Aggregation} & \textbf{Epochs} &  \textbf{LR} & \textbf{k\textsubscript{max}} & \textbf{HU\textsubscript{state}} & \textbf{HU\textsubscript{out}} \\ 
			\hline
			M1(I) & sum & 700 & $4 \times 10^{-3}$ & 5 & 30 & 50 \\
			M1(II) & sum & 1500 & $2 \times 10^{-3}$ & 6 & 100 & 60 \\
			M1(III) & sum & 2000 & $1 \times 10^{-5}$ & 6 & 100 & 60 \\
			\hline
			M2(I) & avg & 500 & $2 \times 10^{-3}$ & 3 & 20 & 50 \\
			M2(II) & avg & 1000 & $1 \times 10^{-3}$ & 4 & 40 & 60 \\
			\hline
			M3(I) & avg & 500 & $2 \times 10^{-3}$ & 6 & 20 & 50 \\
			M3(II) & sum & 500 & $2 \times 10^{-3}$ & 6 & 20 & 50 \\
			M3(III)* & avg & 500 & $2 \times 10^{-3}$ & 6 & 20 & 50\\
			M3(IV)* & sum & 500 & $2 \times 10^{-3}$ & 6 & 20 & 50\\
			\hline	
		\end{tabular}
	}
	\caption{\small{Different module configurations for QM9 are identified by the module number M1, M2 or M3 introduced in Subsection \ref{Method:models}, and by a sequential version number (I, II, ...). Hyperparameters correspond to: neighbor aggregation function (Aggregation), training epochs (Epochs), initial learning rate (LR), maximum state convergence iterations (k\textsubscript{max}), hidden units of the state network (HU\textsubscript{state}), and hidden units of the output network (HU\textsubscript{out}). M3 versions marked with * were trained with class weights to balance the supervisions.}}
	\label{table:hyperparameters}
\end{table}

\subsection{Evaluation}
\label{Experiments:evaluation}
\noindent
The evaluation of generation performance is twofold. On the one hand, the metrics for unconditional generation introduced in \cite{GraphVAE} are used to measure the validity, uniqueness and novelty of the generated graphs. On the other hand, the distributions of the chemical properties of the compounds can be compared to those measured on the test set, assessing the model's capability of reproducing the characteristics of QM9 compounds. Both evaluations are carried out on batches of 10,000 generated graphs.
\\
Let $Gen$ be the set\footnote{More precisely, here we are using the multiset, an extension of the standard set which can contain multiple copies of the same instances.} of generated compounds, $Val \subseteq Gen$ be the subset of chemically valid compounds, and $QM9$ be the set of molecules in the dataset. Validity is calculated as the fraction of chemically valid molecules over the total generated molecules: $Validity = |Val| / |Gen|$. Uniqueness is the fraction of unique molecules among the valid ones: $Uniqueness = |uniq(Val)| / |Val|$, where $uniq$ is a function that takes in input a multiset and returns the corresponding set, from which the duplicates are removed. Novelty is the fraction of unique molecules which do not match any QM9 compound: $Novelty = (|uniq(Val)|-|uniq(Val) \cap QM9|) / |uniq(Val)|$. We also define an additional measure, that combines the three previous metrics and accounts for the fraction of valid, unique and novel molecules over the total generated ones: $VUN = Validity \times Uniqueness \times Novelty$ \footnote{The goals of optimizing validity, uniqueness or novelty are usually in contrast with each other. For instance, improving novelty often comes at the cost of decreasing validity. For this reason, we decided to introduce the new metric $VUN$, which, by combining the three measures, may provide a more global view on the performance of a model.}.
\\
The chemical properties include the molecular weight of each compound, the logarithmic octanol/water partition coefficient (logP) \cite{Crippen_1999}, and the quantitative estimate of drug--likeness (QED) score \cite{QED}. The logP coefficient quantifies the solubility of a molecule in polar or non--polar solvents, while the QED score assesses the drug--likeness of a compound, summarizing in a single measure the following chemical descriptors: polar surface area, molecular weight, logP, number of rotatable bonds, numbers of hydrogen bond donors and acceptors, number of aromatic rings, potential structural issues. 
\\
The validity as well as the chemical properties of each compound are assessed with the RDKit package \cite{rdkit}. In order to determine the uniqueness and novelty of a molecule, we resorted to the graph isomorphism function of the NetworkX package \cite{networkx}. 

\subsection{Results and Discussion}
\label{Experiments:results}
\begin{figure}[!htb]
	\centering
	\begin{minipage}{0.49\textwidth}
		\includegraphics[width=\linewidth]{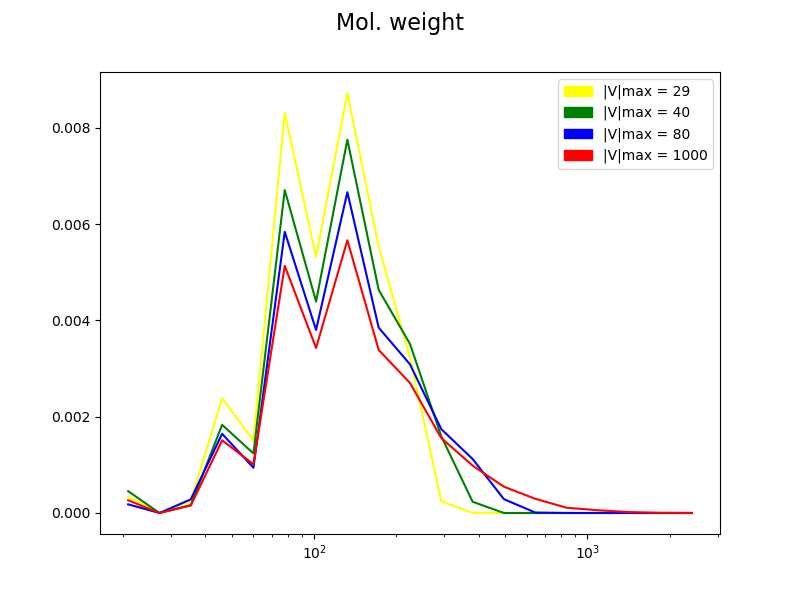}
	\end{minipage}\hfill
	\begin{minipage}{0.49\textwidth}
		\includegraphics[width=\linewidth]{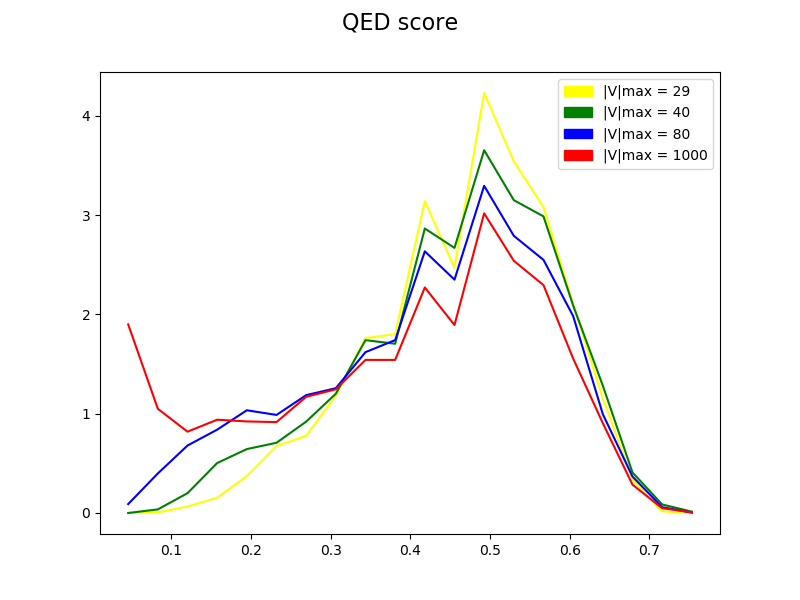}
	\end{minipage}\hfill
	\caption{\small{Logarithm of the molecular weight (left) and QED (right) distributions of generated graphs with different values of $|V|_{max}$. It can be observed how higher thresholds imply the generation of heavier compounds, with lower QED.}}
	\label{figure:vmax}
\end{figure}

\noindent
The first experiment, which was carried out on the QM9 dataset, consisted in a study of the role played in the algorithm by the hyperparameter  $|V|_{max}$ which controls the maximum number of nodes in a generated graph. In principle, our model, being trained step by step, can extend its generation procedure for an arbitrary number of iterations, until it stops itself on a complete molecule. This feature could be exploited to extend the generation domain to molecules which are larger than those seen during training, while retaining the same generic patterns. Using M1(I), M2(I) and M3(I), defined in Table \ref{table:hyperparameters}, we explored different thresholds for the maximum number of generated nodes $|V|_{max}$. The natural value for this dataset is $|V|_{max} = 29$, which corresponds to the largest graph size in the training set. As described in Section \ref{Method:algorithm}, the generation procedure stops when the number of vertices reaches $|V|_{max}$. This means that any graph still incomplete at that point will not correspond to a valid molecule. Intuitively, raising $|V|_{max}$ will increase the amount of valid generated compounds. Even if this is confirmed by the results reported in Table \ref{table:vmax}, the additional valid molecules, being heavier than average, alter the property distributions of the batch. Moreover, as shown in Fig. \ref{figure:vmax}, their QED is below average. Falling in a region of low to very--low drug--likeness, these compounds are not useful in the scope of generating new potential drugs. These considerations suggested to keep $|V|_{max}=29$ for the subsequent experiments.

\begin{table}
	\centering
	\scalebox{0.75}{
		\begin{tabular}{ccccc}
			\hline
			\textbf{Max size} & \textbf{Validity} & \textbf{Uniqueness} & \textbf{Avg. QED} & \textbf{Avg. Mol. Wt.} \\
			\hline
			29 & 0.491 & 0.813 & \textbf{0.448} & \textbf{124.6} \\
			40 & 0.593 & 0.845 & 0.438 & 144.7 \\
			80 & 0.688 & 0.866 & 0.408 & 172.9 \\
			1000 & \textbf{0.781} & \textbf{0.879} & 0.366 & 231.3 \\
			\hline
		\end{tabular}
	}
	\caption{\small{Higher values of $|V|_{max}$, on generation batches from the same model setup, produce more valid and unique compounds. The divergence of average QED and molecular weight from the values taken on the validation set (0.478 and 127.5, respectively), however, suggests that the best configuration is $|V|_{max}=29$.}}
	\label{table:vmax}
\end{table}

\noindent
Starting from the baseline configuration C1 (see Table \ref{table:setups}), in which the modules are optimized separately and not on the evaluation metrics chosen for our task, we explored the hyperparameter space in search of a better configuration\footnote{A systematic search on a grid of configurations was computationally infeasible. Moreover, since the generative models are evaluated with antagonist metrics, it is impossible to optimize the configuration for all of them. Thus, we have heuristically selected the most promising solutions and reported those experiments which, in our opinion, are the most interesting.}. The first step consisted in increasing $k_{max}$ and the number of hidden units in the first two modules, in order for them to better capture complex molecular patterns during training. Using this new configuration (M1(II) and M2(II)), we explored different setups for M3. In particular, to establish the ideal neighbor aggregation method, M3(I) was compared to M3(II). Then, the same comparison was repeated on M3(III) and M3(IV), which were trained with class weighted supervisions\footnote{The error on each pattern is multiplied by the inverse of the  prior of its target class. In this way, the GNN will tend to produce a balanced output over all the classes. At test time, the output is re--multiplied by the vector of prior class probabilities, to restore this important piece of information.}. This latter measure was meant to decrease the learning bias from the very unbalanced prior class probability distribution, which could prevent the model from learning the chemical rules (roughly, $97\%$ of the examples belong to the $disconnected$ class, while the other $3\%$ are distributed over the three bond types). The results of these configurations can be observed in Table \ref{table:linker}.

\begin{table}
	\scalebox{0.75}{
		\begin{tabular}{ccccccc}
			\hline
			\textbf{M3 Module} & \textbf{M3 Agg.} & \textbf{M3 Wts.} & \textbf{Validity} & \textbf{Uniqueness} & \textbf{Avg. QED} & \textbf{Avg. Mol. Wt.} \\
			\hline
			M3(I) & avg & no & \textbf{0.511} & 0.888 & \textbf{0.461} & 134.8 \\
			M3(II) & sum & no & 0.507 & 0.887 & 0.460 & 135.1 \\
			M3(III) & avg & yes & 0.476 & \textbf{0.892} & 0.459 & \textbf{134.2} \\
			M3(IV) & sum & yes & 0.499 & 0.888 & 0.460 & 134.3 \\
			\hline
		\end{tabular}
	}
	\caption{\small{Alternative setups for M3 on QM9. Balancing weights bring no advantage on model performance. The two aggregation functions show equivalent results.}}
	\label{table:linker}
\end{table}

\noindent
This balancing strategy for module M3 did not bring advantages, as it is shown in Table \ref{table:linker}. This suggests that the GNN can deal with the unbalanced distribution, and efforts to improve the generation performance should focus on other parameters. Besides, the two neighbor aggregation methods appear to be equivalent. A new version of the node generation module, M1(III) was also trained, increasing the number of training epochs and decreasing the initial learning rate (see Table \ref{table:hyperparameters}), in order to stabilize the learning process and avoid early suboptimal solutions. The relevant setups of our model, produced in these experiments, are summarized in Table \ref{table:setups}.

\begin{table}
	\scalebox{0.75}{
		\begin{tabular}{cccccccc}
			\hline
			\textbf{Config.} & \textbf{M1} & \textbf{M2} & \textbf{M3} & \textbf{Validity} & \textbf{Uniqueness} & \textbf{Avg. QED} & \textbf{Avg. Mol. Wt.} \\
			\hline
			C1 & M1(I) & M2(I) & M3(I) & 0.491 & 0.813 & 0.448 & \textbf{124.6} \\
			C2 & M1(II) & M2(II) & M3(I) & 0.511 & \textbf{0.888} & \textbf{0.461} & 134.8 \\
			C3 & M1(III) & M2(II) & M3(II) & \textbf{0.668} & 0.340 & 0.404 & 75.3 \\
			\hline
		\end{tabular}
	}
	\caption{\small{Summary of the best configurations determined by preliminary experiments on QM9. C3 produces more valid molecules, while the highest QED is obtained by C2. C1 has the closest average molecular weight to the validation set reference (127.5).}}
	\label{table:setups}
\end{table}

\noindent
Table \ref{table:comparison} compares the results achieved by the most interesting configurations of the proposed MG\textsuperscript{2}N\textsuperscript{2} to various baselines, including the the state of the art for unconditional generation on QM9 (see Subsection \ref{Experiments:evaluation} for the metrics). In particular, we compared to: ChemVAE \cite{GomezBombarelli2018}, which is based on SMILES strings, and represents a good baseline which does not exploit a graph representation; GrammarVAE \cite{GrammarVAE} which is also based on SMILES, and exploits the grammar of this string representation of molecules; MolGAN \cite{MolGAN}, which is the best sequential model on this dataset; and GraphVAE \cite{GraphVAE}, which is a very competitive (VAE based) method; MPGVAE \cite{MPGVAE}, a VAE approach in which both the encoder and the decoder are Message Passing Neural Networks\cite{MPNN}. The average values and standard deviations of the chemical descriptors are compared to the equivalent measures from the test set. As for the MolGAN approach \cite{MolGAN}, our model does not include the computation of likelihood, nor is it optimized for the global reconstruction of the training examples, as VAEs do \footnote{VAEs learn to reconstruct the training examples as closely as possible. The reconstruction penalty is calculated on a global basis, as the Kullback-Leibler divergence between the example graph and its reconstructed version. As the KL-divergence cannot be directly optimized, due to the presence of intractable terms, VAEs optimize the Evidence Lower BOund (ELBO) of these terms, which provides a valuable method to enforce a good global reconstruction.}. The lack of an explicit global reconstruction penalty is one of the reasons for the very high novelty of the material produced by MG\textsuperscript{2}N\textsuperscript{2}: the model is not forced to perfectly reconstruct the molecules on a global basis, but it is forced to correctly reconstruct the local parts of the graph. This approach is expected to preserve a certain degree of validity while encouraging the model to explore more different molecular patterns. Though GraphVAE and MolGAN have higher validity, our model outperforms both of them in terms of uniqueness of the compounds. MPGVAE almost reaches the validity shown by MolGAN, while also achieving good uniqueness, and novelty, and outperforming the other approaches. This advantage is mainly due to the message passing steps performed on the graph in the encoding/decoding pipeline. The aggregated VUN score shows that MG\textsuperscript{2}N\textsuperscript{2} generates the highest percentage of molecules which are valid, unique, and novel at the same time. Notice that, differently to all of the baselines, our method explicitly generates the hydrogen atoms, and all of the hydrogens are required to have been explicitly generated to mark a molecule as valid. This difference is one of the factors determining the lower performance of our approach on the validity metric.

\begin{table}
	\scalebox{0.75}{
		\begin{tabular}{cccccccc}
			\hline
			\textbf{Model} & \textbf{Valid} & \textbf{Unique} & \textbf{Novel} & \textbf{VUN} & \textbf{Avg. QED} & \textbf{Avg. logP} & \textbf{Avg. Mol. Wt.} \\
			\hline
			ChemVAE & 0.103 & 0.675 & 0.900 & 0.063 & - & - & - \\
			MPGVAE & 0.910 & 0.680 & 0.540 & 0.334 & - & - & - \\ 
			GrammarVAE & 0.602 & 0.093 & 0.809 & 0.045 & - & - & - \\
			GraphVAE & 0.557 & 0.760 & 0.616 & 0.261 & - & - & - \\
			MolGAN & \textbf{0.981} & 0.104 & 0.942 & 0.096 & - & - & - \\
			\hline
			Ours(C2) & 0.511 & \textbf{0.888} & \textbf{1.000} & \textbf{0.454} & 0.461 (0.116) & 0.272 (1.336) & 134.8 (45.7) \\
			Ours(C3) & 0.668 & 0.340 & \textbf{1.000} & 0.227 & 0.404 (0.088) & 0.238 (1.093) & 75.3 (52.8) \\
			\hline
			Test & - & - & - & - & 0.482 (0.096) & 0.270 (1.325) & 127.3 (7.6) \\
			\hline
		\end{tabular}
	}
	\caption{\small{Validity, Uniqueness, and Novelty of generated compounds assessing the quality of our models and the baselines on the QM9 dataset. The average values of chemical descriptors (Molecular Weight, logP, and QED) are compared to the same quantities measured over the test set. Standard deviations are reported between parentheses. Metrics for GrammarVAE, ChemVAE  and GraphVAE are taken from the GraphVAE article \cite{GraphVAE}. The performance of MolGAN \cite{MolGAN} and MPGVAE \cite{MPGVAE} are taken from their respective papers.}}
	\label{table:comparison}
\end{table}

\begin{figure}[!htb]
	\centering
	\begin{minipage}{0.49\textwidth}
		\includegraphics[width=\linewidth]{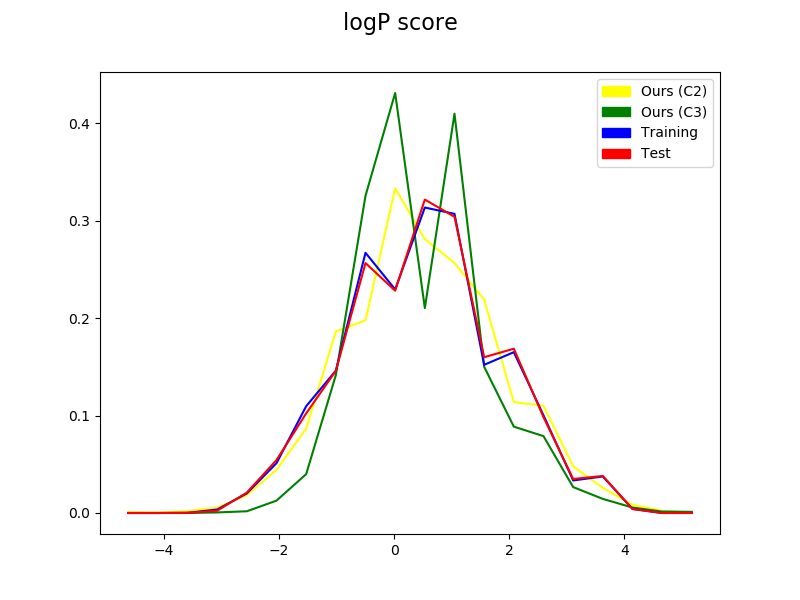}
	\end{minipage}\hfill
	\begin{minipage}{0.49\textwidth}
		\includegraphics[width=\linewidth]{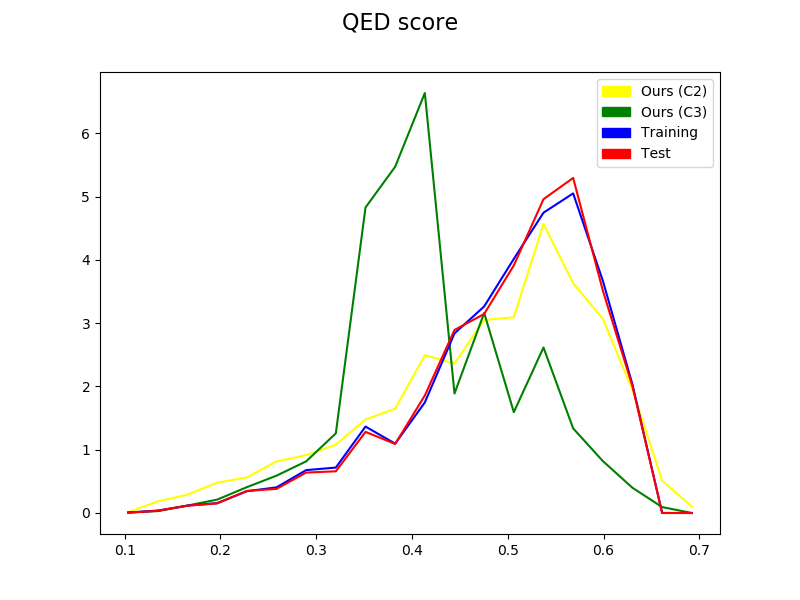}
	\end{minipage}\hfill
	\caption{\small{logP (left) and QED (right) distributions of generated graphs and training/test molecules. It can be observed how well C2 has generalized the chemical characteristics of the compounds seen during training.}}
	\label{figure:descriptors}
\end{figure}

\begin{figure}[!htb]
	\centering
	\begin{minipage}{0.98\textwidth}
		\includegraphics[width=\linewidth]{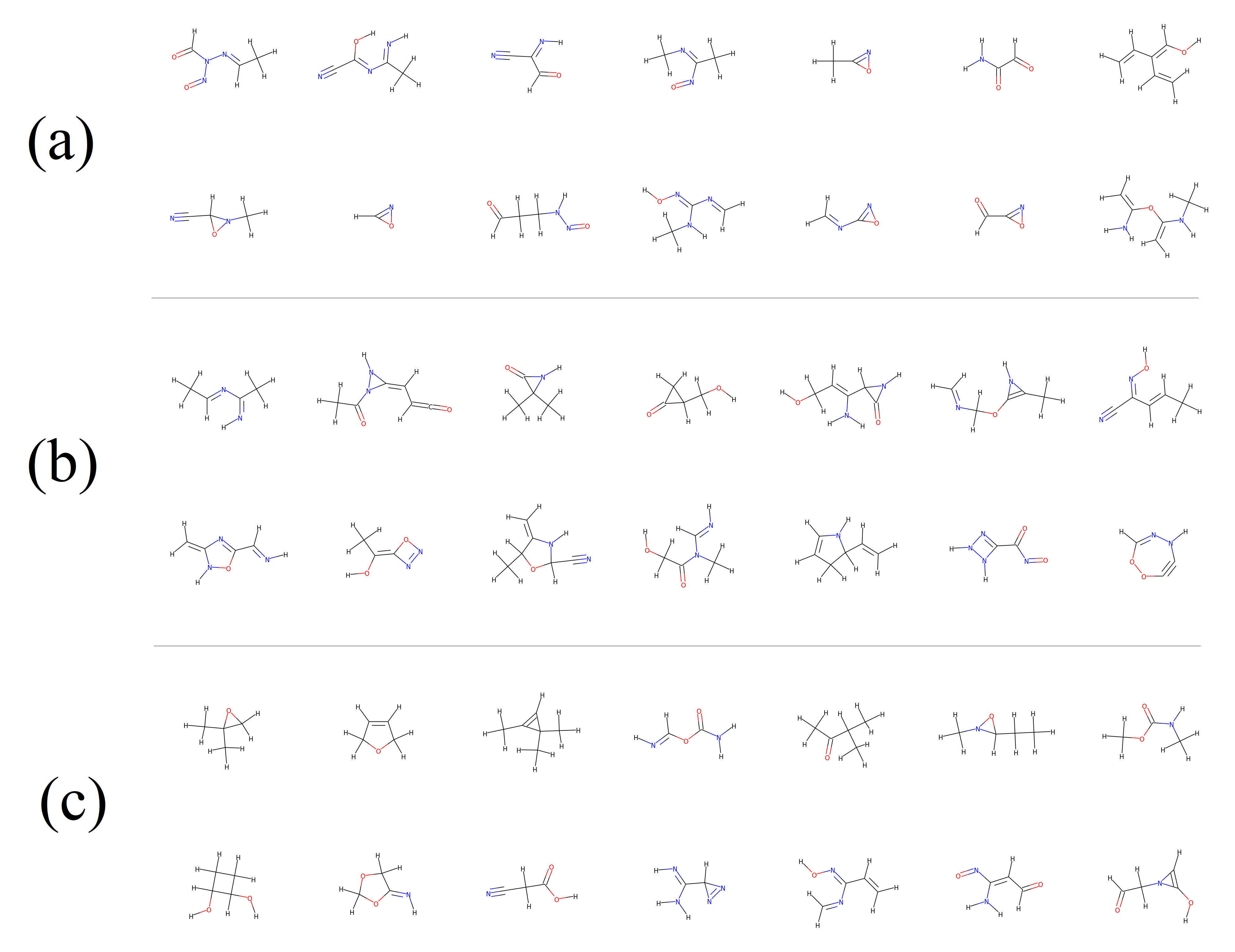}
	\end{minipage}\hfill
	\caption{\small{Grid representation of random samples of 14 valid molecular graphs generated with configuration C2 (a), 14 valid molecular graphs generated with configuration C3 (b), and 14 molecular graphs from the QM9 test set (c).}}
	\label{figure:grid}
\end{figure}

\noindent
To further assess the chemical similarity between the generated material and the test set of molecules from QM9, we plotted the distributions of the chemical descriptors, which can be observed in Fig. \ref{figure:descriptors}. For a qualitative visual comparison, showing the similarity between test set graphs and generated graphs we extracted some valid molecules at random from each set and plotted their structural formulas with RDKit (see Fig. \ref{figure:grid}).
\\
While achieving an acceptable logP distribution, configuration C3 fails to reproduce the QED distribution of the test set. Configuration C2, instead, generates compounds which have very similar logP and QED distributions with respect to those of the test set. This is due to the further optimization carried out on C3: while achieving the goal of building more valid compounds, it actually went in contrast with the other objectives of generating unique, novel molecules with QM9--like properties. The learning parameters proved to have a role in determining the properties of the model, as we can see by comparing C2 and C3. C2 can be considered as our best model configuration for QM9.

\noindent
To further assess the performance of our model, a set of experiments was carried out on the Zinc dataset. An optimization procedure analogous to the one described in Subsection \ref{Experiments:setup} for QM9 allowed to set up the three modules independently. The hyperparameters were then optimized according to the VUN aggregated score. The best model configuration, which was used for the final experiment, is reported in table \ref{table:zinc_hyperparameters}.

\begin{table}
	\centering
	\scalebox{0.75}{
		\begin{tabular}{ccccccc}
			\hline
			\textbf{Module} & \textbf{Aggregation} & \textbf{Epochs} &  \textbf{LR} & \textbf{k\textsubscript{max}} & \textbf{HU\textsubscript{state}} & \textbf{HU\textsubscript{out}} \\ 
			\hline
			M1(Zinc) & sum & 2000 & $10^{-3}$ & 6 & 150 & 80 \\
			\hline
			M2(Zinc) & avg & 1000 & $10^{-3}$ & 4 & 50 & 70 \\
			\hline
			M3(Zinc) & avg & 500 & $2 \times 10^{-3}$ & 6 & 20 & 50 \\
			\hline	
		\end{tabular}
	}
	\caption{\small{Module configurations used in the Zinc experiment, identified by the module number M1, M2 or M3 introduced in Subsection \ref{Method:models}. Hyperparameters correspond to: neighbor aggregation function (Aggregation), training epochs (Epochs), initial learning rate (LR), maximum state convergence iterations (k\textsubscript{max}), hidden units of the state network (HU\textsubscript{state}), and hidden units of the output network (HU\textsubscript{out}).}}
	\label{table:zinc_hyperparameters}
\end{table}

\noindent
To determine the best network parameters for the experiments on Zinc, we started from the best model configuration on QM9, and performed a small grid search in their vicinity, as a more extensive parameter search would have had an infeasible time and computational cost. It can be noticed that modules M1 and M2 required an increased number of parameters to converge, with respect to the QM9 case. This is due to the larger size of the Zinc molecular graphs (up to 38 heavy atoms) compared to the QM9 ones (up to 9 heavy atoms), and to the larger number of node and edge types. The larger size of Zinc molecules also implies a longer generation sequence (on wider graphs), which added to the larger number of examples, and to the larger number of network parameters, multiplies the time and memory burden of each experiment. For this reason, we limited the experimentation on Zinc to the comparison with other models in the literature. Table \ref{table:zinc_comparison} reports the performance of competitive models which were tested for unconditional generation on Zinc, according to the Validity, Uniqueness, and Novelty metrics defined in the GraphVAE paper \cite{GraphVAE}, and to the VUN aggregated score defined in Subsection \ref{Experiments:evaluation}. In particular, our model is compared to GraphVAE, ChemVAE \cite{GomezBombarelli2018}, GrammarVAE \cite{GrammarVAE}, and the state of the art approach CGVAE \cite{CGVAE}. The large difference in performance between CGVAE and the other methods is largely justified by the different approach to the problem. In CGVAE, molecules are kekulized during the preprocessing stage, thus aromatic bonds are reduced to either single or double bonds. The other hard chemical laws, like valence rules, that would invalidate the generated molecule if violated, are enforced as hard constraints, preventing the possibility of generating invalid molecules \cite{CGVAE}. In all the other reported methods, including ours, these rules are learned by the model. Contrarily to what happended on QM9, in this case our model outperforms the standard VAE baselines thanks to the higher validity. This suggests that, as the number of atom types grows, the importance of generating the atoms sequentially and re-examining the output graph at each step, also grows.

\begin{table}
	\centering
	\scalebox{0.75}{
		\begin{tabular}{ccccc}
			\hline
			\textbf{Model} & \textbf{Valid} & \textbf{Unique} & \textbf{Novel} & \textbf{VUN} \\
			\hline
			GrammarVAE\cite{GrammarVAE} & 0.310 & 0.108 & \textbf{1.000} & 0.033 \\
			ChemVAE\cite{GomezBombarelli2018} & 0.170 & 0.310 & 0.980 & 0.052\\
			GraphVAE\cite{GraphVAE} & 0.140 & 0.316 & \textbf{1.000} & 0.044 \\
			CGVAE\cite{CGVAE} & \textbf{1.000} & \textbf{0.998} & \textbf{1.000} & \textbf{0.998}\\
			\hline
			Ours & 0.753 & 0.107 & \textbf{1.000} & 0.081\\
			\hline
		\end{tabular}
	}
	\caption{\small{Validity, Uniqueness, and Novelty of generated compounds assessing the quality of our model and the baselines on the Zinc dataset. The performance of the other models are taken from the CGVAE article\cite{CGVAE}}}.
	\label{table:zinc_comparison}
\end{table}

\section{Conclusions}
\label{Conclusions}
\noindent
In this article, we introduced a generative model for molecular graphs: Molecule Generative Graph Neural Network (MG\textsuperscript{2}N\textsuperscript{2}), and a sequential generation algorithm we devised for this purpose. The novelty of our approach consists in exploiting the capabilities of Graph Neural Networks to natively process graph--structured data. This allows to use the graph output of the previous step as the network input, which represents an advantage with respect to other sequential methods, that mainly rely on the sequence of previous decisions, rather than on the graph they produce. The modularity of our model implies an easier, less resource demanding, learning process. 
\\
In line with all the other sequential methods, and contrarily to VAEs, the generation process is easily interpretable: the steps in which errors occur, or in which specific atoms and bonds are created, can be readily identified in the generation sequence. This feature is very important as it simplifies any process of improvement or repurposing of the model.
\\
The model was tested on a benchmark generation task over the QM9 dataset. The distributions of the chemical descriptors retraced those measured on the held out test set. The quality of generated graphs proved to be very high, allowing our model to outperform very competitive baselines. The same performance level was observed also on the Zinc dataset, when comparing our model to similar approaches.
\\
Future work will focus on generalizing the model to other molecular graph generation problems, and on extending the present approach to conditional generation. A conditional generation model could be implemented by concatenating a vector of desired properties to the input of each module. The comparison with a completely different approach, like CGVAE, which simplifies the generation problem by enforcing chemical rules as hard constraints, suggests that a constrained, or fragment--based, version of our model could improve the performance on datasets of larger molecules, like Zinc. Moreover, studying a theoretical mathematical formulation of sequential generation is also an important matter of future research.

\section*{Acknowledgements}
This research did not receive any specific grant from funding agencies in the public, commercial, or not-for-profit sectors.

\bibliographystyle{ieeetr}
\bibliography{references}

\begin{thebibliography}{10}

\bibitem{ErdosRenyi}
P.~{Erd\"H{o}s} and A.~{R{\'e}nyi}, ``On random graphs {I},'' {\em
  Publicationes Mathematicae Debrecen}, vol.~6, pp.~290--297, 1959.

\bibitem{BarabasiAlbert}
R.~{Albert} and A.-L. {Barab\'asi}, ``Statistical mechanics of complex
  networks,'' {\em Rev. Mod. Phys.}, vol.~74, pp.~47--97, 2002.

\bibitem{WattsStrogatz}
D.~J. {Watts} and S.~H. {Strogatz}, ``Collective dynamics of 'small-world'
  networks,'' {\em Nature}, vol.~393, no.~6684, pp.~440--442, 1998.

\bibitem{VAE_original}
D.~P. {Kingma} and M.~{Welling}, ``Auto--encoding variational {B}ayes,'' in
  {\em Proceedings of the 2nd International Conference on Learning
  Representations (ICLR)}, 2014.

\bibitem{VGAE}
T.~N. {Kipf} and M.~{Welling}, ``Variational graph auto-encoders,'' in {\em
  NeurIPS Bayesian Deep Learning Workshop}.

\bibitem{Graphite}
A.~{Grover}, A.~{Zweig}, and S.~{Ermon}, ``Graphite: Iterative generative
  modeling of graphs,'' in {\em International Conference on Machine Learning},
  pp.~2434--2444, 2019.

\bibitem{GAN_original}
I.~{Goodfellow}, J.~{Pouget-Abadie}, M.~{Mirza}, B.~{Xu}, D.~{Warde-Farley},
  S.~{Ozair}, A.~{Courville}, and Y.~{Bengio}, ``Generative adversarial nets,''
  in {\em Advances in neural information processing systems}, pp.~2672--2680,
  2014.

\bibitem{GraphGAN}
H.~{Wang}, J.~{Wang}, J.~{Wang}, M.~{Zhao}, W.~{Zhang}, F.~{Zhang}, X.~{Xie},
  and M.~{Guo}, ``Graph{GAN}: Graph representation learning with generative
  adversarial nets,'' in {\em Thirty-second AAAI conference on artificial
  intelligence}, pp.~2508--2515, 2018.

\bibitem{NetGAN}
A.~{Bojchevski}, O.~{Shchur}, D.~{Z{\"u}gner}, and S.~{G{\"u}nnemann},
  ``Net{GAN}: Generating graphs via random walks,'' in {\em International
  Conference on Machine Learning}, pp.~610--619, 2018.

\bibitem{ConstrainedGraphGAN}
L.~{Di Liello}, P.~{Ardino}, J.~{Gobbi}, P.~{Morettin}, S.~{Teso}, and
  A.~{Passerini}, ``Efficient generation of structured objects with constrained
  adversarial networks,'' {\em Advances in Neural Information Processing
  Systems}, vol.~33, 2020.

\bibitem{GraphRNN}
J.~{You}, R.~{Ying}, X.~{Ren}, W.~L. {Hamilton}, and J.~{Leskovec},
  ``Graph{RNN}: Generating realistic graphs with deep auto--regressive
  models,'' in {\em International Conference on Machine Learning},
  pp.~5708--5717, 2018.

\bibitem{GRAN}
R.~{Liao}, Y.~{Li}, Y.~{Song}, S.~{Wang}, W.~{Hamilton}, D.~K. {Duvenaud},
  R.~{Urtasun}, and R.~{Zemel}, ``Efficient graph generation with graph
  recurrent attention networks,'' in {\em Advances in Neural Information
  Processing Systems}, pp.~4255--4265, 2019.

\bibitem{edge_based_sequential_graph_generation}
D.~{Bacciu}, A.~{Micheli}, and M.~{Podda}, ``Edge--based sequential graph
  generation with recurrent neural networks,'' {\em Neurocomputing}, vol.~416,
  pp.~177--189, 2020.

\bibitem{GNN_Model}
F.~{Scarselli}, M.~{Gori}, A.~C. {Tsoi}, M.~{Hagenbuchner}, and
  G.~{Monfardini}, ``The graph neural network model,'' {\em IEEE Transactions
  on Neural Networks}, vol.~20, no.~1, pp.~61--80, 2009.

\bibitem{GraphNets}
P.~W. {Battaglia}, J.~B. {Hamrick}, V.~{Bapst}, A.~{Sanchez-Gonzalez},
  V.~{Zambaldi}, M.~{Malinowski}, A.~{Tacchetti}, D.~{Raposo}, A.~{Santoro},
  R.~{Faulkner}, C.~{Gulcehre}, F.~{Song}, A.~{Ballard}, J.~{Gilmer},
  G.~{Dahl}, A.~{Vaswani}, K.~{Allen}, C.~{Nash}, V.~{Langston}, C.~{Dyer},
  N.~{Heess}, D.~{Wierstra}, P.~{Kohli}, M.~{Botvinick}, O.~{Vinyals}, Y.~{Li},
  and R.~{Pascanu}, ``Relational inductive biases, deep learning, and graph
  networks,'' 2018.

\bibitem{GraphSAGE}
W.~Hamilton, Z.~Ying, and J.~Leskovec, ``Inductive representation learning on
  large graphs,'' in {\em Advances in neural information processing systems},
  pp.~1024--1034, 2017.

\bibitem{GGSNN}
Y.~{Li}, D.~{Tarlow}, M.~{Brockschmidt}, and R.~{Zemel}, ``Gated graph sequence
  neural networks,'' 2015.

\bibitem{MPNN}
J.~{Gilmer}, S.~S. {Schoenholz}, P.~F. {Riley}, O.~{Vinyals}, and G.~E. {Dahl},
  ``Neural message passing for quantum chemistry,'' in {\em Proceedings of the
  34th International Conference on Machine Learning}, vol.~70, pp.~1263--1272,
  2017.

\bibitem{GCN_spectral_Bruna}
J.~{Bruna}, W.~{Zaremba}, A.~{Szlam}, and Y.~{LeCun}, ``Spectral networks and
  deep locally connected networks on graphs,'' in {\em 2nd International
  Conference on Learning Representations, ICLR 2014}, 2014.

\bibitem{GCN_standard}
T.~N. {Kipf} and M.~{Welling}, ``Semi--supervised classification with graph
  convolutional networks,'' in {\em 5th International Conference on Learning
  Representations, ICLR 2017}.

\bibitem{GCN_spectral_Defferrard}
M.~{Defferrard}, X.~{Bresson}, and P.~{Vandergheynst}, ``Convolutional neural
  networks on graphs with fast localized spectral filtering,'' in {\em Advances
  in neural information processing systems}, pp.~3844--3852, 2016.

\bibitem{DeepGMG}
Y.~{Li}, O.~{Vinyals}, C.~{Dyer}, R.~{Pascanu}, and P.~{Battaglia}, ``Learning
  deep generative models of graphs,'' 2018.

\bibitem{drug_discovery_costs}
O.~J. {Wouters}, M.~{McKee}, and J.~{Luyten}, ``{Estimated Research and
  Development Investment Needed to Bring a New Medicine to Market,
  2009-2018},'' {\em JAMA}, vol.~323, no.~9, pp.~844--853, 2020.

\bibitem{SMILES}
D.~{Weininger}, A.~{Weininger}, and J.~L. {Weininger}, ``Smiles. 2. {A}lgorithm
  for generation of unique {SMILES} notation,'' {\em Journal of chemical
  information and computer sciences}, vol.~29, no.~2, pp.~97--101, 1989.

\bibitem{GomezBombarelli2018}
R.~{G{\'o}mez-Bombarelli}, J.~N. {Wei}, D.~{Duvenaud}, J.~M.
  {Hern{\'a}ndez-Lobato}, B.~{S{\'a}nchez-Lengeling}, D.~{Sheberla},
  J.~{Aguilera-Iparraguirre}, T.~D. {Hirzel}, R.~P. {Adams}, and
  A.~{Aspuru-Guzik}, ``Automatic chemical design using a data--driven
  continuous representation of molecules,'' {\em ACS central science}, vol.~4,
  no.~2, pp.~268--276, 2018.

\bibitem{GrammarVAE}
M.~J. {Kusner}, B.~{Paige}, and J.~M. {Hern{\'a}ndez-Lobato}, ``Grammar
  variational autoencoder,'' in {\em International Conference on Machine
  Learning}, pp.~1945--1954, 2017.

\bibitem{SyntaxVAE}
H.~{Dai}, Y.~{Tian}, B.~{Dai}, S.~{Skiena}, and L.~{Song}, ``Syntax--directed
  variational autoencoder for structured data,'' in {\em 6th International
  Conference on Learning Representations (ICLR)}, 2018.

\bibitem{GraphVAE}
M.~{Simonovsky} and N.~{Komodakis}, ``Graph{VAE}: Towards generation of small
  graphs using variational autoencoders,'' in {\em International Conference on
  Artificial Neural Networks}, pp.~412--422, Springer, 2018.

\bibitem{JTVAE}
W.~{Jin}, R.~{Barzilay}, and T.~{Jaakkola}, ``Junction tree variational
  autoencoder for molecular graph generation,'' in {\em International
  Conference on Machine Learning}, pp.~2323--2332, 2018.

\bibitem{hierarchical_molecule_generation}
W.~Jin, R.~Barzilay, and T.~Jaakkola, ``Hierarchical generation of molecular
  graphs using structural motifs,'' 2020.

\bibitem{CCGVAE}
D.~{Rigoni}, N.~{Navarin}, and A.~{Sperduti}, ``Conditional constrained graph
  variational autoencoders for molecule design,'' 2020.

\bibitem{GraphFlow_autoregressive}
C.~Shi, M.~Xu, Z.~Zhu, W.~Zhang, M.~Zhang, and J.~Tang, ``Graph{AF}: a
  flow--based autoregressive model for molecular graph generation,'' in {\em
  7th International Conference on Learning Representations (ICLR)}, 2019.

\bibitem{MolGAN}
N.~{De Cao} and T.~{Kipf}, ``Mol{GAN}: An implicit generative model for small
  molecular graphs,'' 2018.

\bibitem{GCPN}
J.~{You}, B.~{Liu}, Z.~{Ying}, V.~{Pande}, and J.~{Leskovec}, ``Graph
  convolutional policy network for goal--directed molecular graph generation,''
  in {\em Advances in neural information processing systems}, pp.~6410--6421,
  2018.

\bibitem{gumbel_softmax}
E.~{Jang}, S.~{Gu}, and B.~{Poole}, ``Categorical reparameterization with
  gumbel-softmax,'' in {\em 5th International Conference on Learning
  Representations (ICLR)}, 2017.

\bibitem{GNN_universal_approximation}
F.~Scarselli, M.~Gori, A.~C. Tsoi, M.~Hagenbuchner, and G.~Monfardini,
  ``Computational capabilities of graph neural networks,'' {\em IEEE
  Transactions on Neural Networks}, vol.~20, no.~1, pp.~81--102, 2008.

\bibitem{GNN_Implementation}
A.~{Rossi}, M.~{Tiezzi}, G.~M. {Dimitri}, M.~{Bianchini}, M.~{Maggini}, and
  F.~{Scarselli}, ``Inductive--transductive learning with graph neural
  networks,'' in {\em Artificial Neural Networks in Pattern Recognition. ANNPR
  2018, Lecture Notes in Computer Science, vol 11081} (T.~E. Pancioni~L.,
  Schwenker~F., ed.), pp.~201--212, Springer, Cham, 2018.

\bibitem{backpropagation_through_time}
P.~J. {Werbos}, ``Backpropagation through time: what it does and how to do
  it,'' {\em Proceedings of the IEEE}, vol.~78, no.~10, pp.~1550--1560, 1990.

\bibitem{gumbel_original}
E.~J. {Gumbel}, ``Statistical theory of extreme values and some practical
  applications: a series of lectures,'' 1954.

\bibitem{gumbel_a_star}
C.~J. {Maddison}, D.~{Tarlow}, and T.~{Minka}, ``A* sampling,'' in {\em
  Advances in Neural Information Processing Systems}, pp.~3086--3094, 2014.

\bibitem{freeman_betweenness_centrality}
L.~C. {Freeman}, ``A set of measures of centrality based on betweenness,'' {\em
  Sociometry}, vol.~40, no.~1, pp.~35--41, 1977.

\bibitem{dataset_QM9}
R.~{Ramakrishnan}, P.~O. {Dral}, M.~{Rupp}, and O.~A. {von Lilienfeld},
  ``Quantum chemistry structures and properties of 134 kilo molecules,'' {\em
  Scientific Data}, vol.~1, 2014.

\bibitem{dataset_Zinc}
J.~J. {Irwin} and B.~K. {Shoichet}, ``Zinc- a free database of commercially
  available compounds for virtual screening,'' {\em Journal of chemical
  information and modeling}, vol.~45, no.~1, pp.~177--182, 2005.

\bibitem{dataset_GDB17}
L.~{Ruddigkeit}, R.~{van Deursen}, L.~C. {Blum}, and J.-L. {Reymond},
  ``Enumeration of 166 billion organic small molecules in the chemical universe
  database {GDB}--17,'' {\em J. Chem. Inf. Model.}, vol.~52, no.~11,
  pp.~2864--2875, 2012.

\bibitem{GNN_Code}
N.~{Pancino}, A.~{Rossi}, G.~{Ciano}, G.~{Giacomini}, S.~{Bonechi},
  P.~{Andreini}, F.~{Scarselli}, M.~{Bianchini}, and p.~y. Pietro~{Bongini},
  booktitle={28th European Symposium on Artificial Neural Networks,
  Computational Intelligence and Machine Learning (online event)}, ``Graph
  neural networks for the prediction of protein–protein interfaces,''

\bibitem{tensorflow}
M.~{Abadi}, A.~{Agarwal}, P.~{Barham}, E.~{Brevdo}, Z.~{Chen}, C.~{Citro},
  G.~S. {Corrado}, A.~{Davis}, J.~{Dean}, M.~{Devin}, S.~{Ghemawat},
  I.~{Goodfellow}, A.~{Harp}, G.~{Irving}, M.~{Isard}, Y.~{Jia},
  R.~{Jozefowicz}, L.~{Kaiser}, M.~{Kudlur}, J.~{Levenberg}, D.~{Man\'{e}},
  R.~{Monga}, S.~{Moore}, D.~{Murray}, C.~{Olah}, M.~{Schuster}, J.~{Shlens},
  B.~{Steiner}, I.~{Sutskever}, K.~{Talwar}, P.~{Tucker}, V.~{Vanhoucke},
  V.~{Vasudevan}, F.~{Vi\'{e}gas}, O.~{Vinyals}, P.~{Warden}, M.~{Wattenberg},
  M.~{Wicke}, Y.~{Yu}, and X.~{Zheng}, ``{TensorFlow}: Large--scale machine
  learning on heterogeneous systems,'' 2015.
\newblock Software available from tensorflow.org.

\bibitem{Adam}
D.~P. Kingma and J.~Ba, ``Adam: A method for stochastic optimization,'' 2014.

\bibitem{Crippen_1999}
S.~A. {Wildman} and G.~M. {Crippen}, ``Prediction of physicochemical parameters
  by atomic contributions,'' {\em Journal of Chemical Information and Computer
  Sciences}, vol.~39, no.~5, pp.~868--873, 1999.

\bibitem{QED}
G.~R. {Bickerton}, G.~V. {Paolini}, J.~{Besnard}, S.~{Muresan}, and A.~L.
  {Hopkins}, ``Quantifying the chemical beauty of drugs,'' {\em Nature
  chemistry}, vol.~4, no.~2, pp.~90--98, 2012.

\bibitem{rdkit}
``{RDK}it: Open--source cheminformatics,'' 2016.

\bibitem{networkx}
A.~{Hagberg}, P.~{Swart}, and D.~S. {Chult}, ``Exploring network structure,
  dynamics, and function using network{X},'' in {\em In Proceedings of the 7th
  Python in Science Conference (SciPy)} (J.~M. G.~Varoquaux, T.~Vaught, ed.),
  pp.~11--15, 2008.

\bibitem{MPGVAE}
D.~Flam-Shepherd, T.~Wu, and A.~Aspuru-Guzik, ``Graph deconvolutional
  generation,'' 2020.

\bibitem{CGVAE}
Q.~{Liu}, M.~{Allamanis}, M.~{Brockschmidt}, and A.~{Gaunt}, ``Constrained
  graph variational autoencoders for molecule design,'' {\em Advances in Neural
  Information Processing Systems}, vol.~31, pp.~7795--7804, 2018.

\end{thebibliography}
\end{document}